\documentclass[acmtog,nonacm]{acmart}

\renewcommand\footnotetextcopyrightpermission[1]{}
\usepackage{adjustbox}
\usepackage{multirow}

\usepackage{booktabs} % For formal tables
\usepackage{colortbl} % For row coloring
\usepackage{placeins} % Provides \FloatBarrier to constrain float placement

\usepackage[ruled]{algorithm2e} % For algorithms

\SetAlFnt{\small}
\SetAlCapFnt{\small}
\SetAlCapNameFnt{\small}
\SetAlCapHSkip{0pt}

\acmJournal{TOG}
\begin{document}
% Title portion
\title{HandFlow: Fully Generative 4D Hand Recovery with Flow Matching}

% 作者信息（最小档：仅 name + institution + city/postcode/country）
% 机构：①HKUST(GZ)  ②Google
% Mingxi Xu 与 Bowen Duan 为共同第一作者，Yutao Yue 为通讯作者
\author{Mingxi Xu}
\authornote{These authors contributed equally to this work.}
\affiliation{%
  \institution{The Hong Kong University of Science and Technology (Guangzhou)}
  \city{Guangzhou}
  \postcode{511400}
  \country{China}}

\author{Bowen Duan}
\authornotemark[1]
\affiliation{%
  \institution{The Hong Kong University of Science and Technology (Guangzhou)}
  \city{Guangzhou}
  \postcode{511400}
  \country{China}}

\author{Yi Gu}
\affiliation{%
  \institution{The Hong Kong University of Science and Technology (Guangzhou)}
  \city{Guangzhou}
  \postcode{511400}
  \country{China}}

\author{Zhengyang Shen}
\affiliation{%
  \institution{Google}
  \city{Mountain View}
  \state{CA}
  \country{USA}}

\author{Renjing Xu}
\affiliation{%
  \institution{The Hong Kong University of Science and Technology (Guangzhou)}
  \city{Guangzhou}
  \postcode{511400}
  \country{China}}

\author{Yutao Yue}
\authornote{Corresponding author.}
\affiliation{%
  \institution{The Hong Kong University of Science and Technology (Guangzhou)}
  \city{Guangzhou}
  \postcode{511400}
  \country{China}}

\renewcommand\shortauthors{Xu et al.}

\begin{abstract}
Accurate monocular 4D hand reconstruction remains challenging. Per-frame discriminative regressors lack temporal context and often produce jittery predictions. Temporal models improve consistency by aggregating information across frames, but they are typically deterministic regressors, making them vulnerable to ambiguous observations caused by occlusion and motion blur.
Generative modeling offers a natural alternative by learning a prior over plausible hand motion sequences, enabling coherent hand-state recovery when visual evidence is incomplete or unreliable.

Motivated by this observation, we present HandFlow, a fully generative flow-matching framework for temporally coherent 3D hand pose and shape estimation from monocular video. Given visual and skeletal observations, HandFlow denoises an entire temporal window of MANO parameters through a single ODE integration. To support this, we use a Flux-style dual-stream transformer that attends across the full sequence to capture long-range dependencies without autoregressive decoding, and a confidence-aware continuous masking mechanism that blends observed features with learnable mask tokens to handle noisy or missing observations. Experiments on DexYCB and HOT3D show that HandFlow achieves state-of-the-art performance, with particularly large gains in world-space accuracy and temporal smoothness. It reduces world-space pose error by over 30\% compared with the strongest baseline and achieves the lowest acceleration error among all evaluated methods, while remaining competitive in per-frame pose accuracy. Moreover, on a single GPU HandFlow reconstructs a 150-frame sequence at 47 fps, about $12\times$ faster than the fastest prior video-based method, with reconstruction itself accounting for only a small fraction of the end-to-end latency. Our model and code are publicly available at \url{https://mxxu00.github.io/HandFlow/}.

\end{abstract}

%
% The code below should be generated by the tool at
% http://dl.acm.org/ccs.cfm
% Please copy and paste the code instead of the example below.
%


%
% End generated code
%


\begin{teaserfigure}
\centering
\includegraphics[width=\textwidth]{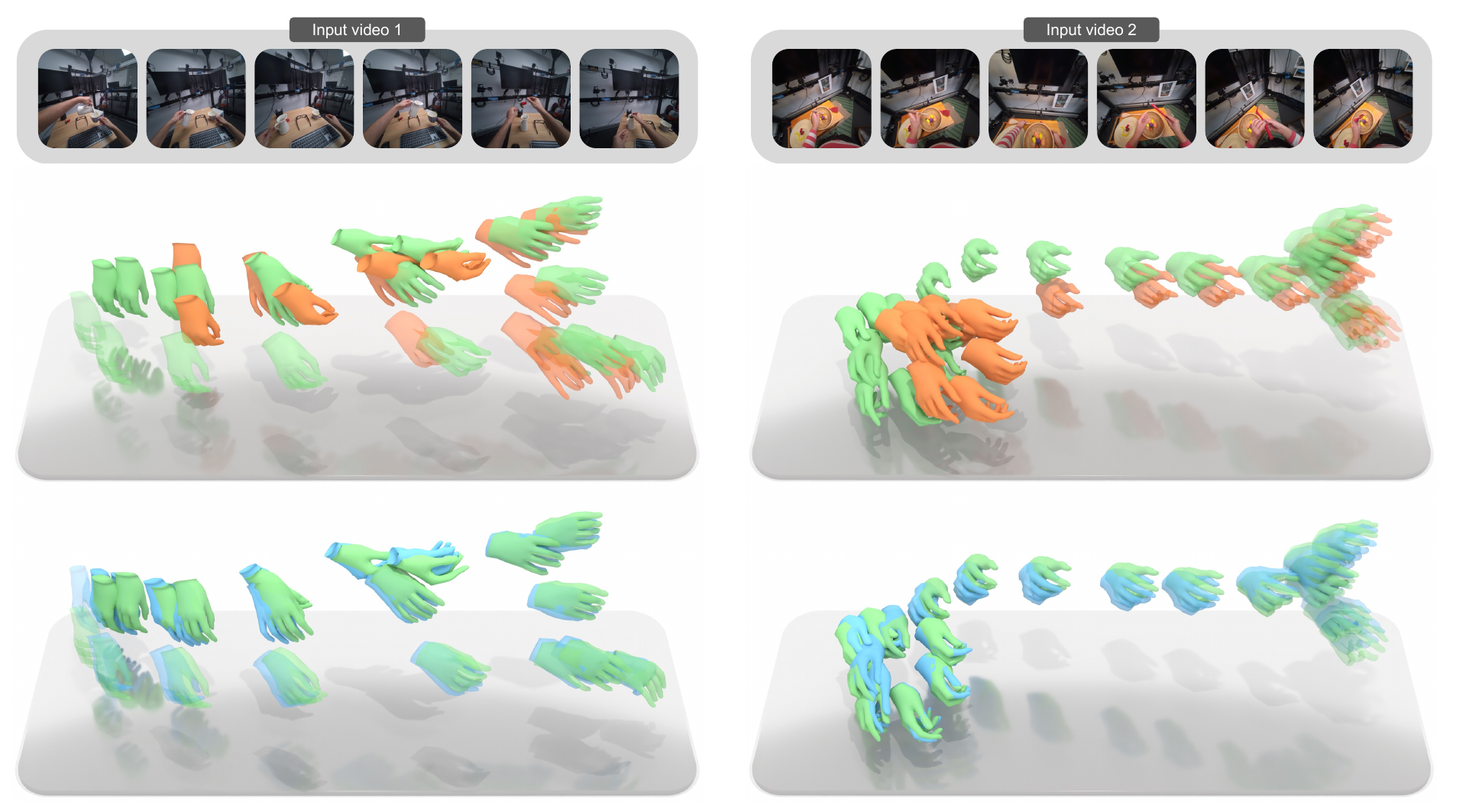}
\caption{
% Given monocular RGB videos, HandFlow reconstructs temporally coherent 4D hand motion through fully generative flow matching. Compared with HaWoR~\cite{zhang_hawor_2025}, which shows noticeable temporal drift and hand-pose misalignment, HandFlow recovers smoother and better aligned hand trajectories across challenging sequences with rapid motion and large viewpoint changes, while reconstructing each window over an order of magnitude faster than HaWoR. Green denotes the ground-truth hand motion, orange denotes HaWoR, and blue denotes HandFlow.
Given monocular RGB videos, HandFlow reconstructs temporally coherent 4D hand motion through a fully generative flow-matching framework. Compared with HaWoR~\cite{zhang_hawor_2025} (orange), which shows noticeable temporal drift and hand-pose misalignment, HandFlow (blue) recovers smoother, more accurately aligned hand trajectories across challenging sequences involving rapid motion and large viewpoint changes. It also reconstructs each window more than an order of magnitude faster than HaWoR. Ground-truth hand motion is shown in green.
}
\Description[Teaser figure comparing HandFlow with HaWoR]{Side-by-side comparison of 4D hand reconstruction on three challenging sequences. Green: ground-truth motion, orange: HaWoR output with visible temporal drift, blue: HandFlow output with smooth and well-aligned trajectories.}
\label{fig:teaser} 
\end{teaserfigure}

% 去掉首页页脚自动生成的 "Authors' addresses" 重复块
% 必须置于所有 \author/\affiliation 之后：那些命令会填充内部宏 \@authorsaddresses，
% acmart 据此宏是否为空决定是否输出地址脚注；清空它即跳过整个脚注。
\makeatletter
\let\@authorsaddresses\@empty
\makeatother

\maketitle

\section{Introduction}
\label{sec:intro}
Accurate monocular 4D hand reconstruction aims to recover temporally coherent 3D hand meshes from a single RGB video, with broad applications in AR/VR~\cite{wang2020rgb2hands}, robotics~\cite{christen2024synh2r, pan2025spider}, and human-computer interaction~\cite{liu2022hoi4d}. Despite substantial progress in hand mesh recovery, stable and accurate reconstruction over long videos remains challenging. This difficulty largely stems from the inherent ambiguity of monocular observations. Hands often move rapidly, self-occlude, interact with objects, and partially or fully leave the camera view. Under such conditions, a single image, or even a short video clip, can correspond to multiple plausible 3D hand states.

Most existing methods formulate hand reconstruction as a deterministic regression problem. 
Per-frame methods~\cite{pavlakos_reconstructing_2024, potamias_wilor_2025, chen_handos_2025} estimate each frame independently and can perform well when the hand is clearly visible. However, without temporal reasoning, their predictions often fluctuate across frames. Temporal methods~\cite{fu_deformer_2023, zhang_hawor_2025, yu_dyn-hamr_2025, ye_predicting_2026} alleviate this jitter by aggregating information over time. 
Nevertheless, they typically still produce a single point estimate for a given input video. When the visual evidence is ambiguous, such a deterministic mapping cannot represent the multimodal posterior over plausible hand motions, often resulting in inaccurate, over-smoothed, or kinematically implausible predictions, as illustrated in Figure~\ref{fig:teaser}.

This motivates a generative view of monocular 4D hand reconstruction. Rather than treating the video-to-motion mapping as a deterministic regression problem, we cast reconstruction as conditional generation in the MANO parameter space~\cite{romero_embodied_2017}. The key idea is to recover a trajectory that explains the available visual evidence while staying on the manifold of plausible hand motion.

Based on this perspective, we propose \textbf{HandFlow}, a rectified-flow~\cite{liu_flow_2023} framework for temporally coherent 4D hand reconstruction from monocular video. Rather than regressing a single output, HandFlow learns the distribution of plausible MANO motion sequences conditioned on the observed visual and skeletal evidence, and samples a coherent trajectory through a single ODE integration. Two structural challenges shape its design. The first, modeling long-range dependencies across the full temporal window, is addressed by a Flux-style dual-stream transformer that jointly attends to hand parameters and visual-skeletal conditions over the full sequence without autoregressive decoding. The second, unreliable observations under occlusion and motion blur, is addressed by \emph{cmask}, a confidence-aware continuous masking strategy that interpolates between observed features and learnable mask tokens according to per-frame detection confidence. An overlapping-window scheme with velocity blending further extends reconstruction to long videos while preserving efficiency.

In summary, our main contributions are as follows:
\begin{itemize}
  \item We formulate monocular 4D hand reconstruction as conditional generation in the MANO parameter space and introduce \textbf{HandFlow}, a fully generative flow-matching framework for reconstructing temporally coherent motion sequences from monocular video.

  \item We design a dual-stream transformer that jointly models hand parameters and visual-skeletal conditions over a full temporal window without autoregressive decoding.

  \item We introduce confidence-aware continuous masking (\emph{cmask}) that interpolates between observed conditional features and learnable mask tokens according to per-frame detection confidence, improving robustness to occlusion and unreliable observations.

  \item We propose velocity-blended overlapping-window inference, whose parallelizable window decomposition delivers high-throughput reconstruction with a tunable memory-throughput budget.
  
\end{itemize}

\section{Related Work}
\label{sec:related_work}
\noindent
\textbf{Single-image 3D hand reconstruction.}
Early methods estimated 3D hand pose from 2D keypoints via inverse kinematics, but were limited by keypoint detection accuracy.
The introduction of the MANO parametric hand model~\cite{romero_embodied_2017} enabled end-to-end training by providing a differentiable hand representation.
Recent transformer-based methods have significantly advanced single-image reconstruction.
HaMeR~\cite{pavlakos_reconstructing_2024} scales up both data and model size with a ViT-Huge backbone, achieving robust in-the-wild performance.
WiLoR~\cite{potamias_wilor_2025} further improves localization and reconstruction through a convolutional hand detector combined with a transformer regressor.
HandOS~\cite{chen_handos_2025} unifies detection and reconstruction in a one-stage framework.
While these methods achieve strong per-frame accuracy, they process each image independently and produce temporally inconsistent outputs when applied to video.

\noindent
\textbf{Video-based 4D hand reconstruction.}
Several methods have been proposed to leverage temporal information for more consistent hand motion estimation.
Deformer~\cite{fu_deformer_2023} introduces a dynamic fusion module that warps predictions from adjacent frames to support the current frame, combined with a maxMSE loss that focuses on error-prone joints.
TCMR~\cite{choi_beyond_2021} addresses temporal inconsistency in human body reconstruction by balancing past and future frame information.
HaWoR~\cite{zhang_hawor_2025} reconstructs world-space hand motion from egocentric video by decoupling camera-space hand estimation from SLAM-based camera trajectory recovery.
Dyn-HaMR~\cite{yu_dyn-hamr_2025} tackles the challenging setting of dynamic cameras through a multi-stage optimization pipeline.
HaPTIC~\cite{ye_predicting_2026} predicts 4D hand trajectories using cross-view and global cross-attention layers on top of an image-based transformer.
Despite modeling temporal dependencies, these methods share a fundamental limitation: they are \emph{deterministic regressors} that map observations to a single output, making them vulnerable to ambiguous inputs where multiple hand states are equally plausible.

\noindent
\textbf{Generative models for pose and motion.}
Diffusion models have demonstrated remarkable success in generating diverse and high-quality outputs across image~\cite{xu_magicanimate_2024} and motion domains.
In human motion generation, MDM~\cite{tevet_human_2023} and similar methods apply diffusion to generate motion sequences from text or action conditioning.
More recently, UniHand~\cite{sun_unihand_2026} proposes a unified diffusion-based framework that formulates both estimation and generation as conditional motion synthesis, using a VAE to compress structured signals into a shared latent space.
HandDiff~\cite{cheng_handdiff_2024} applies diffusion to 3D hand pose estimation from image-point clouds but operates on single frames.
MaskHand~\cite{saleem_maskhand_2025} uses masked modeling with VQ-MANO tokens for single-image reconstruction with confidence-guided sampling.
Most related to our formulation, FMPose3D~\cite{wang_fmpose3d_2026} applies flow matching to monocular 3D pose estimation, but targets single-frame body pose from 2D keypoints, whereas HandFlow generates temporally coherent 4D hand motion in the MANO parameter space from multimodal visual-skeletal conditions.
DuoMo~\cite{wang_duomo_2026} reconstructs world-space human motion with a dual diffusion design; HandFlow instead recovers world-space hand trajectories through a single flow-matching pass followed by SLAM-based camera pose estimation.
In contrast, HandFlow operates directly on raw MANO parameters without latent compression (preserving full representational capacity) and uses a rectified-flow formulation~\cite{liu_flow_2023} that needs far fewer integration steps than diffusion sampling.

\section{Method}
\label{sec:method}
\begin{figure*}[t] 
\centering
\includegraphics[width=0.98\linewidth]{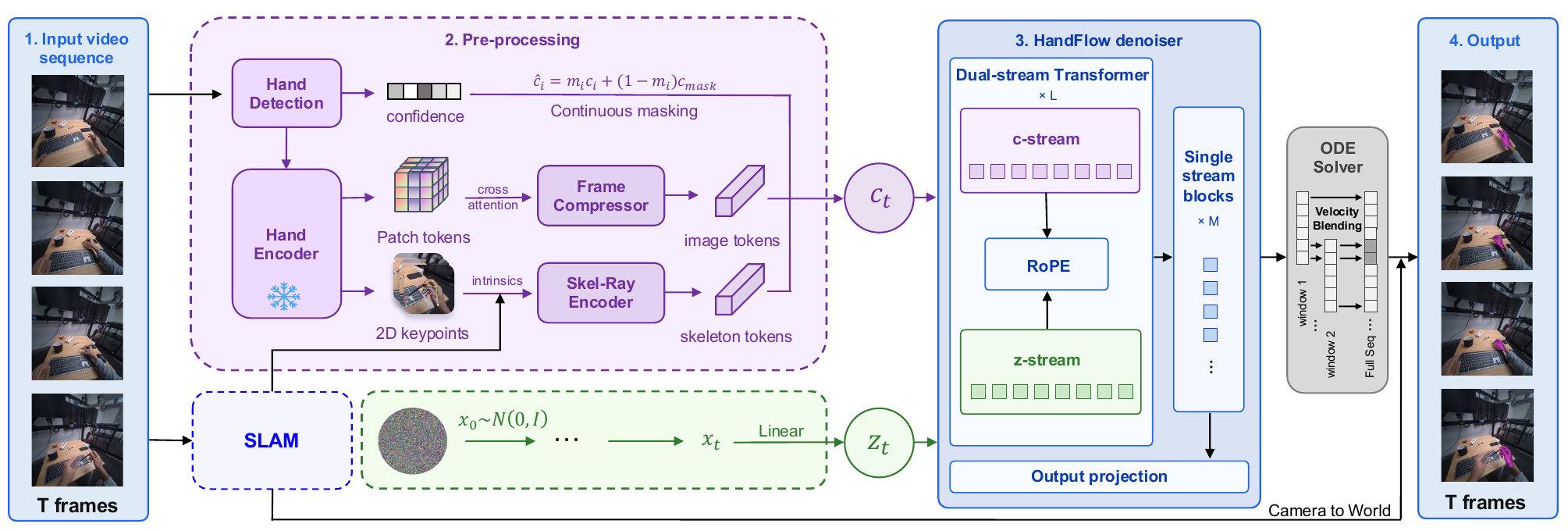}
\caption{
Overview of HandFlow. A frozen HaMeR~\cite{pavlakos_reconstructing_2024} frontend extracts visual tokens, 2D skeletons, and confidence, which are compressed into condition tokens with confidence-aware continuous masking. A dual-stream flow-matching transformer then denoises MANO pose, translation, and shape tokens from noise over the full temporal window; for long videos, overlapping windows are merged by velocity blending during ODE integration.
}
\Description[HandFlow pipeline overview]{Pipeline of HandFlow. A frozen HaMeR frontend extracts hand crops, visual tokens, 2D skeletons, and confidence from an input video. A conditioning stage compresses the visual tokens, encodes the 2D keypoints as camera rays, and applies confidence-aware continuous masking. A dual-stream flow-matching transformer denoises MANO pose, translation, and shape tokens from Gaussian noise over the full temporal window, and overlapping windows are merged by velocity blending for long-video inference.}
\label{fig:pipeline} 
\end{figure*}
Given a monocular video clip of $T$ frames, our goal is to recover a temporally coherent sequence of 3D hand meshes in the MANO parameter space~\cite{romero_embodied_2017} via rectified flow~\cite{liu_flow_2023}.
As illustrated in Figure~\ref{fig:pipeline}, our pipeline consists of three stages: (1)~extracting visual and skeletal conditions from the input video, with a continuous masking mechanism to handle unreliable observations, (2)~denoising a packed MANO parameter sequence from noise to signal with a dual-stream transformer through a single ODE integration, and (3)~a composite training objective combining flow matching with geometric and temporal constraints.

Each hand state is represented by shared shape coefficients $\boldsymbol{\beta} \in \mathbb{R}^{10}$, per-frame joint poses $\boldsymbol{\theta}_i \in \mathbb{R}^{48}$ in axis-angle format, and global translations $\boldsymbol{\tau}_i \in \mathbb{R}^{3}$. We pack the parameters within a temporal window into a single flat vector and apply per-component standardization by subtracting the dataset mean and dividing by the standard deviation independently for each dimension:
\begin{equation}
\bar{\mathbf{x}}_1 = \operatorname{normalize}\left(
[\boldsymbol{\beta} \mid \boldsymbol{\theta}_1, \dots, \boldsymbol{\theta}_T \mid \boldsymbol{\tau}_1, \dots, \boldsymbol{\tau}_T]
\right) \in \mathbb{R}^{d}, \quad d = 10 + 51T.
\label{eq:flow_target}
\end{equation}
All flow-matching operations are performed in this normalized space. We de-normalize the predictions before computing geometric auxiliary losses and applying MANO forward kinematics.

\subsection{Rectified Flow Formulation}
\label{sec:flow_matching}

We adopt the rectified flow framework~\cite{liu_flow_2023, lipman_flow_2023}, a special case of flow matching that connects noise and data with a straight (linear) interpolation path, chosen for its few-step sampling efficiency.
At training time, we sample a timestep $t \sim p(t)$ and construct an intermediate state between Gaussian noise and the normalized MANO target:
\begin{equation}
  \bar{\mathbf{x}}_t = (1 - t)\,\bar{\mathbf{x}}_0 + t\,\bar{\mathbf{x}}_1,
  \label{eq:interpolation}
\end{equation}
where $\bar{\mathbf{x}}_0 \sim \mathcal{N}(\mathbf{0}, \mathbf{I})$ denotes standard Gaussian noise. Under this linear path, the target velocity is given by $\mathbf{v}^* = \bar{\mathbf{x}}_1 - \bar{\mathbf{x}}_0$. The denoiser $f_\phi$ is trained to predict this velocity $\mathbf{v}^*$ from the noisy state $\bar{\mathbf{x}}_t$, timestep $t$, and conditioning signal $\mathbf{c}$:

\begin{equation}
  \mathcal{L}_\text{fm} = \left\| f_\phi(\bar{\mathbf{x}}_t, t, \mathbf{c}) - \mathbf{v}^* \right\|_2^2.
  \label{eq:loss_fm}
\end{equation}
Following~\cite{blackforestlabs_flux_2024}, we sample $t$ from a logit-normal distribution and apply a time-shift mapping, concentrating training density in the middle timesteps to improve sampling efficiency for sequence-level generation.

At inference, we solve the ODE from $t{=}0$ to $t{=}1$ using Euler integration with 3 steps (a stable operating point on the accuracy plateau; see Appendix~\ref{sec:app_ode_sweep}), starting from random noise and arriving at a denoised MANO parameter sequence $\hat{\bar{\mathbf{x}}}_1$, which is then de-normalized to recover the final predictions.

\subsection{Condition Representation}
\label{sec:condition_representation}
The conditioning signal $\mathbf{c}$ is constructed from two complementary modalities: dense visual features and sparse skeletal cues. Both are extracted frame-wise using a frozen HaMeR detector~\cite{pavlakos_reconstructing_2024}.
Given an input image, HaMeR localizes the hand, crops a $256 \times 256$ hand region, and predicts 21 2D keypoints together with detection confidence scores.

\noindent
\textbf{Visual features.}
For each cropped hand image, we use the frozen HaMeR ViT-Huge backbone~\cite{pavlakos_reconstructing_2024} to extract patch-level features, which a lightweight frame compressor distills into a single image token $\mathbf{c}^{\mathrm{img}}_i \in \mathbb{R}^{D}$. The compressor consists of 8 cross-attention pooling layers.

\noindent
\textbf{Skeletal features.}
The 2D keypoints predicted by HaMeR provide sparse geometric guidance.
For each of the 21 keypoints, we back-project its image coordinate into a camera-normalized 3D ray using the crop camera intrinsics $(f_x, f_y, c_x, c_y)$:
\begin{equation}
  \mathbf{r}_{i,j} =
  \left[
  \frac{u_{i,j} - c_x}{f_x},\;
  \frac{v_{i,j} - c_y}{f_y},\;
  1
  \right],
\end{equation}
where $(u_{i,j}, v_{i,j})$ represents the 2D coordinates of $j$-th keypoint in the $i$-th frame. Each ray is further encoded with a sinusoidal positional encoding~\cite{mildenhall2021nerf} using 6 frequency bands.
This ray-based representation incorporates camera intrinsics into the skeletal condition, which has been shown to be important for accurate 3D reconstruction~\cite{wang_duomo_2026, baradel2024multi, feng2025physhmr, li2022cliff}.
The encoded rays of all 21 keypoints are flattened and projected by a two-layer MLP to obtain a per-frame skeleton token $\mathbf{c}^{\mathrm{skel}}_i \in \mathbb{R}^{D}$.

\noindent
\textbf{Continuous masking.}
When the hand is heavily occluded, blurred, or outside the camera view, HaMeR typically produces low confidence detections, making both visual and skeletal observations unreliable.
To handle this, we introduce \emph{cmask}, a continuous masking mechanism that interpolates each condition token with a learnable mask token (Eq.~\ref{eq:cmask}).
Let $\mathbf{c}^{k}_i \in \mathbb{R}^{D}$ denote a condition token at frame $i$, where $k \in \{\mathrm{img}, \mathrm{skel}\}$.
The masked condition token is defined as
\begin{equation}
  \hat{\mathbf{c}}^{k}_i
  =
  m_i \mathbf{c}^{k}_i
  +
  (1 - m_i) \mathbf{c}^{k}_{\mathrm{mask}},
  \label{eq:cmask}
\end{equation}
where $m_i \in [0,1]$ is the HaMeR detection confidence and $\mathbf{c}^{k}_{\mathrm{mask}} \in \mathbb{R}^{D}$ is a learnable mask token for modality $k$.
During training, we additionally apply random masking to approximately 20\% of frames as regularization.

The final condition sequence is constructed by concatenating the masked image
and skeleton tokens over all frames:
\begin{equation}
  \mathbf{c}
  =
  [\hat{\mathbf{c}}^{\mathrm{img}}_1,
  \hat{\mathbf{c}}^{\mathrm{skel}}_1,
  \dots,
  \hat{\mathbf{c}}^{\mathrm{img}}_T,
  \hat{\mathbf{c}}^{\mathrm{skel}}_T]
  \in \mathbb{R}^{2T \times D}.
\end{equation}

\subsection{Dual-Stream Transformer}
\label{sec:transformer}
The denoiser adopts a dual-stream architecture inspired by Flux~\cite{blackforestlabs_flux_2024}. It jointly processes latent hand parameters, referred to as the \emph{z-stream}, and conditioning tokens, referred to as the \emph{c-stream}. An overview of the architecture is shown in Figure~\ref{fig:pipeline}.

\noindent
\textbf{Tokenization.}
The \emph{z-stream} tokenizes the noisy state $\bar{\mathbf{x}}_t$ into $1 + 2T$ tokens: one shape token $\mathbf{z}^{\beta}$, $T$ pose tokens $\{\mathbf{z}^{\theta}_i\}_{i=1}^{T}$, and $T$ translation tokens $\{\mathbf{z}^{\tau}_i\}_{i=1}^{T}$.
Each component is projected from its native dimension to the hidden dimension $D$ via a dedicated linear layer.
The \emph{c-stream} takes the $2T$ conditioning tokens described in Section~\ref{sec:condition_representation}.

\noindent
\textbf{Dual-stream blocks.}
Each dual-stream block applies joint attention over the z- and c-streams. The flow timestep $t$ is first encoded with a sinusoidal embedding and then passed through an MLP to produce the adaptive LayerNorm modulation parameters $(\gamma, \beta, \alpha)$. Within each block, QKV projections are computed separately for the \emph{z-stream} and \emph{c-stream}. The resulting queries, keys, and values are then concatenated and fed into a shared attention operation, allowing latent hand tokens to attend to both other latent tokens and the conditioning context.
The attention outputs are subsequently split back into the two streams and processed by stream-specific feed-forward networks.

Rotary positional encoding (RoPE)~\cite{su_roformer_2021} is applied to the queries and keys after QK normalization.
We assign position indices according to temporal alignment: the shape token is assigned position 0; for each frame $i$, the pose token $\mathbf{z}^{\theta}_i$ and translation token $\mathbf{z}^{\tau}_i$ are both assigned position $i$; and the corresponding image and skeleton tokens in the c-stream are also assigned position $i$.

\noindent
\textbf{Single-stream blocks.}
After the dual-stream blocks, the z- and c-tokens are concatenated and further processed by single-stream self-attention blocks following the DiT architecture~\cite{peebles_scalable_2023}, using the same adaptive modulation scheme.

\noindent
\textbf{Output projection.}
Finally, the z-stream tokens are projected back to their native dimensions through dedicated linear layers and unpacked to reconstruct the predicted clean state $\hat{\bar{\mathbf{x}}}_1 \in \mathbb{R}^{d}$.

\subsection{Training Objectives}
\label{sec:training}
The model is trained with a composite loss. The primary term is the flow matching loss $\mathcal{L}_\text{fm}$ (Eq.~\ref{eq:loss_fm}). We further introduce auxiliary supervision on the one-step clean prediction $\hat{\bar{\mathbf{x}}}_1 = \bar{\mathbf{x}}_t + (1 - t)\,\hat{\mathbf{v}}$, obtained by integrating the denoiser's predicted velocity $\hat{\mathbf{v}} = f_\phi(\bar{\mathbf{x}}_t, t, \mathbf{c})$ one step along the flow. We then de-normalize $\hat{\bar{\mathbf{x}}}_1$ back to the original MANO parameter space before computing the losses. Specifically, $\mathcal{L}_{\beta}$ is an $\ell_1$ loss on the shape coefficients; $\mathcal{L}_{\mathrm{vel}}$ and $\mathcal{L}_{\mathrm{acc}}$ are first- and second-order temporal finite-difference losses ($\ell_1$) over the packed MANO parameters (joint poses and global translation jointly), supervising motion smoothness directly in parameter space; $\mathcal{L}_{\mathrm{reproj}}$ is the $\ell_1$ error between the 21 MANO joints of the prediction and of the ground truth, each projected to the $256{\times}256$ crop image plane via the crop camera intrinsics $(f_x, f_y, c_x, c_y)$; and $\mathcal{L}_{\mathrm{j3d}}$ is the $\ell_1$ error, in millimeters, on the absolute global 3D joint positions. All auxiliary losses are scaled by the mean sampled timestep $\bar{t}$. Invalid padding frames are masked out in all loss terms. Loss weights and training hyperparameters are provided in Appendix~\ref{sec:exp_details}.

\subsection{Overlapping Window Inference}
\label{sec:vblend}
For inference on videos longer than $T$ frames, non-overlapping windows leave boundary frames at window edges, where limited temporal context causes abrupt velocity changes that degrade trajectory quality. We therefore adopt overlapping sliding windows.
The full sequence of $N$ frames is divided into windows of length $T$, where
each window shares $o$ frames with the previous one, corresponding to a stride
$s = T - o$.
All windows use identical per-frame initial noise: a global frame $g$ is
assigned the same Gaussian noise vector regardless of which windows contain it.
This ensures that the ODE integration starts from a globally consistent noisy
state.

The critical question is how to merge predictions for frames that appear in multiple windows. A straightforward strategy, which we refer to as \emph{post-hoc averaging}, runs an independent ODE for each window and averages the final denoised states $\hat{\mathbf{x}}_1$ of overlapping frames. However, because different windows follow different ODE trajectories, they can produce inconsistent predictions for the same physical frame in the normalized
parameter space. Averaging these divergent endpoints can introduce unnatural velocity
discontinuities, especially near overlap boundaries (see Appendix~\ref{sec:app_overlap}).

We instead perform \emph{velocity blending}, which fuses predictions
\emph{during} ODE integration by maintaining a single shared state for each
global frame.
Concretely, let global frame $g$ be covered by windows $w_1, \dots, w_k$, and
let $\ell_i$ denote the local frame index of $g$ within window $w_i$.
We assign a center-distance weight
\begin{equation}
  \omega(w_i, \ell_i) =
  \max\!\left(0.01,\; 1 - \frac{|\ell_i - c|}{c}\right),
  \quad c = \frac{T-1}{2},
  \label{eq:vblend_weight}
\end{equation}
which gives larger weights to predictions near the window center, where the
temporal context is more balanced.
At each ODE step, the blended velocity for frame $g$ is computed as
\begin{equation}
  \tilde{\mathbf{v}}_g =
  \frac{\sum_{i=1}^{k} \omega(w_i, \ell_i)\,
    \hat{\mathbf{v}}_{w_i, \ell_i}}
  {\sum_{i=1}^{k} \omega(w_i, \ell_i)},
  \label{eq:vblend}
\end{equation}
The weight $\omega$ governs the per-frame pose and translation velocities; the sequence-shared shape $\boldsymbol{\beta}$ has no per-frame index, so we blend its velocity by an unweighted average over all windows, $\tilde{\mathbf{v}}^{\beta} = \tfrac{1}{k}\sum_{i=1}^{k} \hat{\mathbf{v}}^{\beta}_{w_i}$.
The shared state is updated by
\begin{equation}
  \mathbf{x}_g \leftarrow
  \mathbf{x}_g + \tilde{\mathbf{v}}_g \Delta t,
\end{equation}
where $\Delta t$ denotes the ODE step size.

Since all window velocities are evaluated from the same shared per-frame
state, the resulting trajectory follows a single continuous integration path.
Overlapping frames also benefit from complementary visual-skeletal
observations: when a frame lies near the boundary of one window and suffers
from occlusion or motion blur, an adjacent window may place the same frame
closer to its center and provide a more reliable conditioning context. It is worth noting that all windows at each ODE step can be processed in
parallel as a single batch. Therefore, velocity blending introduces only minor
additional inference overhead while providing substantial improvements in
trajectory continuity, as analyzed in Appendix~\ref{sec:app_overlap}.

\section{Experiments}

\subsection{Datasets}
\label{sec:datasets}
We evaluate our method on two benchmarks spanning both controlled and in-the-wild scenarios. \textbf{DexYCB}~\cite{chao_dexycb_2021} contains 1,000 video sequences of subjects grasping YCB objects, captured with multiple calibrated cameras. We use the standard s0 evaluation protocol and evaluate on the right-hand crops from the default camera view. \textbf{HOT3D}~\cite{banerjee_hot3d_2025} is a recent benchmark with diverse 3D hand annotations for tabletop interactions captured by egocentric multi-view cameras. Its official validation clips have no public ground truth, so, as is common for this benchmark, we hold out a per-participant validation split from the official training set and evaluate on crops from the primary view. Both datasets provide MANO ground-truth annotations. All methods with publicly available code are evaluated end-to-end with their native pipelines. We additionally report earlier baselines~\cite{lin_meshgraphormer_2021, chen_semihandobj_2023, park_handoccnet_2022, chen_s2hand_2021, kocabas_vibe_2020, valassakis_handdgp_2024, duran_hmp_2024}. UniHand~\cite{sun_unihand_2026}, the most closely related generative baseline, is also included; its code and HOT3D split are undocumented (we release both), so we quote its numbers from the original paper. HandFlow nonetheless outperforms it on the large majority of metrics across both benchmarks. Metric definitions are provided in Appendix~\ref{sec:app_metrics}.

\begin{table*}[!ht]
\centering
\caption{Occlusion-robustness evaluation on DexYCB (s0). RA-MPJPE and PA-MPJPE in mm; $AUC_{RA}$ and $AUC_{PA}$ are normalized to $[0,1]$. Best in \textbf{bold}, second best \underline{underlined}. Other baseline results are taken from~\cite{fu_deformer_2023, zhang_hawor_2025, sun_unihand_2026}.}
\label{tab:dexycb_occlusion}
\small
\setlength{\tabcolsep}{4pt}
\begin{adjustbox}{max width=2\columnwidth}
\begin{tabular}{lcccccccccc}
\toprule
& \multicolumn{4}{c}{All} & \multicolumn{2}{c}{Occl. (25\%--50\%)} & \multicolumn{2}{c}{Occl. (50\%--75\%)} & \multicolumn{2}{c}{Occl. (75\%--100\%)} \\
\cmidrule(lr){2-5} \cmidrule(lr){6-7} \cmidrule(lr){8-9} \cmidrule(lr){10-11}
Method & RA-MPJPE $\downarrow$ & $AUC_{RA}$ $\uparrow$ & PA-MPJPE $\downarrow$ & $AUC_{PA}$ $\uparrow$ & PA-MPJPE $\downarrow$ & $AUC_{PA}$ $\uparrow$ & PA-MPJPE $\downarrow$ & $AUC_{PA}$ $\uparrow$ & PA-MPJPE $\downarrow$ & $AUC_{PA}$ $\uparrow$ \\
\midrule
MeshGraphormer    & 16.21 & 0.691 & 6.41 & 0.872 & 6.85 & 0.863 & 7.22 & 0.856 & 7.76  & 0.845 \\
SemiHandObj       & -- & -- & 6.33 & 0.874 & 6.70 & 0.866 & 7.17 & 0.857 & 8.96  & 0.821 \\
HandOccNet        & -- & -- & 5.80 & 0.884 & 6.22 & 0.876 & 6.43 & 0.872 & 7.37  & 0.853 \\
WiLoR             & -- & -- & 5.01 & 0.900 & --   & --    & 5.42 & 0.892 & 5.68  & 0.887 \\
\midrule
S2HAND(V)         & 19.67 & 0.625 & 7.27 & 0.855 & 7.74 & 0.845 & 7.71 & 0.846 & 7.87  & 0.843 \\
VIBE              & 16.95 & 0.675 & 6.43 & 0.871 & 6.72 & 0.865 & 6.84 & 0.864 & 7.06  & 0.858 \\
TCMR              & 16.03 & 0.701 & 6.28 & 0.875 & 6.56 & 0.869 & 6.58 & 0.868 & 6.95  & 0.861 \\
Deformer          & \underline{13.64} & 0.740 & 5.22 & 0.896 & 5.22 & 0.886 & 5.70 & 0.886 & 6.34  & 0.873 \\
HaWoR             & 14.22 & \underline{0.779} & 4.76 & 0.905 & --   & --    & 5.03 & 0.899 & 5.07  & 0.899 \\
UniHand           & -- & -- & \underline{4.08} & \underline{0.918} & \underline{4.22} & \underline{0.913} & \textbf{4.25} & \underline{0.912} & \textbf{4.26} & \textbf{0.912} \\
\midrule
\textbf{HandFlow} & \textbf{8.12} & \textbf{0.839} & \textbf{3.88} & \textbf{0.922} & \textbf{4.21} & \textbf{0.916} & \underline{4.35} & \textbf{0.913} & \underline{4.85} & \underline{0.903} \\
\bottomrule
\end{tabular}
\end{adjustbox}
\end{table*}

\begin{table}[!ht]
\centering
\caption{Comprehensive evaluation on HOT3D (val). World-space accuracy in mm, acceleration error in m/s$^2$, RTE in \%. Image-based methods (HaMeR, WiLoR) use ViPE-SLAM~\cite{huang_vipe_2025} for camera-pose recovery. Best in \textbf{bold}, second best \underline{underlined}. M.S. Opt. is short for multi-stage optimization. Other baseline results are taken from~\cite{zhang_hawor_2025, sun_unihand_2026}.}
\label{tab:hot3d}
\small
\begin{adjustbox}{max width=\columnwidth}
\begin{tabular}{llcccccc}
\toprule
Method & Paradigm & MPJPE $\downarrow$ & PA-MPJPE $\downarrow$ & W-MPJPE $\downarrow$ & WA-MPJPE $\downarrow$ & Accel $\downarrow$ & RTE $\downarrow$ \\
\midrule
HaMeR + SLAM & Regression & - & 9.39 & 156.03 & 43.37 & 19.25 & 4.77 \\
HandDGP + SLAM & Regression & - & 17.88 & 154.30 & 42.93 & 20.17 & 3.18 \\
WiLoR + SLAM & Regression & - & 6.00 & 151.67 & 39.49 & 8.02 & 2.99 \\
HMP + SLAM & Regression & - & 10.51 & 119.41 & 39.46 & 5.50 & 2.79 \\
Dyn-HaMR & M.S. Opt. & 82.08 & 8.22 & 69.11 & 31.01 & \underline{4.77} & \underline{2.44} \\
HaWoR & Regression & \underline{77.97} & 6.32 & 73.88 & 27.37 & 11.57 & 5.74 \\
UniHand & Denoising & - & \textbf{4.76} & \underline{63.97} & \underline{25.24} & 4.93 & - \\
\midrule
\textbf{HandFlow} & \textbf{Denoising} & \textbf{72.00} & \underline{5.49} & \textbf{43.00} & \textbf{16.17} & \textbf{4.18} & \textbf{2.32} \\
\bottomrule
\end{tabular}
\end{adjustbox}
\end{table}

\subsection{Camera-Space Evaluation}
\label{sec:cam_eval}

We assess camera-space pose accuracy and occlusion robustness on DexYCB
(Table~\ref{tab:dexycb_occlusion}); world-space and trajectory metrics are
evaluated on HOT3D in Section~\ref{sec:world_eval}.

\noindent
\textbf{Pose accuracy.}
On the full test set (All), HandFlow is the best method on every metric, with
the largest margin in RA-MPJPE: 8.12\,mm versus 13.64\,mm for the next-best
Deformer, a 41\% relative reduction. The gap holds across the error
distribution: $AUC_{RA}$ reaches 0.839 against HaWoR's 0.779, so the gain is not
concentrated on easy cases.

\begin{figure*}[t]
\centering
\includegraphics[width=0.98\linewidth]{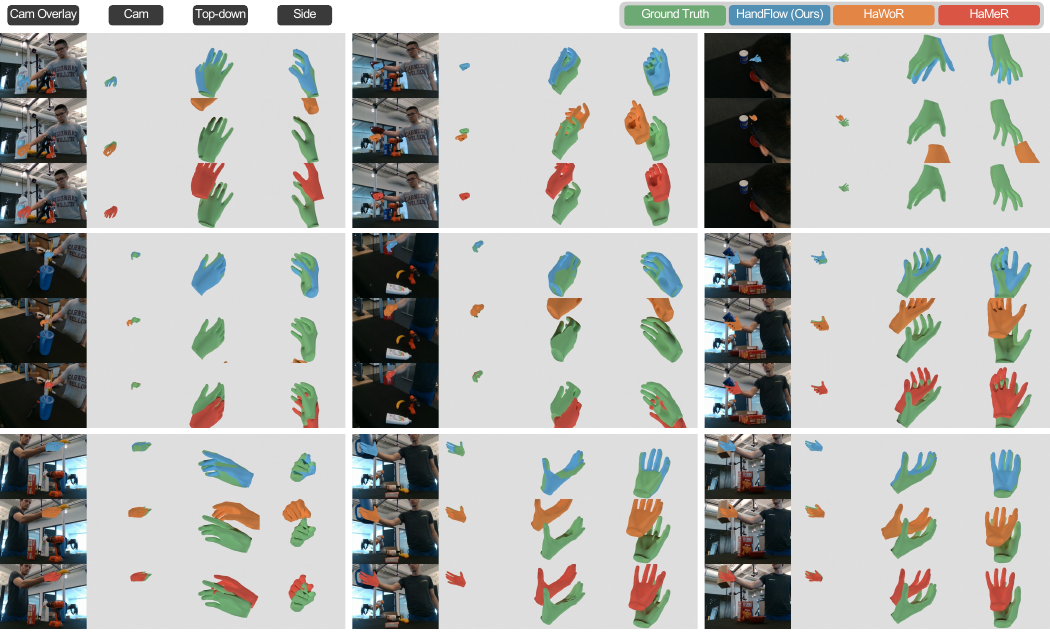}
\caption{Comparison on heavily occluded DexYCB frames: GT vs.\ HaMeR, HaWoR, and HandFlow. Each method is shown in four columns (from left to right: the mesh overlaid on the input, the camera-space mesh alone, a top-down view, and a side view); the last three columns expose depth errors that the overlay hides. Color: green~\textbf{GT}, blue~\textbf{HandFlow}, orange~\textbf{HaWoR}, red~\textbf{HaMeR}.}
\Description[DexYCB occlusion viewpoint comparison]{Heavily occluded DexYCB cases comparing GT with HaMeR, HaWoR, and HandFlow across four columns (the mesh overlaid on the input, the camera-space mesh, top-down, and side), exposing baseline depth errors.}
\label{fig:dexycb-occ}

\vspace{2pt}

\includegraphics[width=0.98\linewidth]{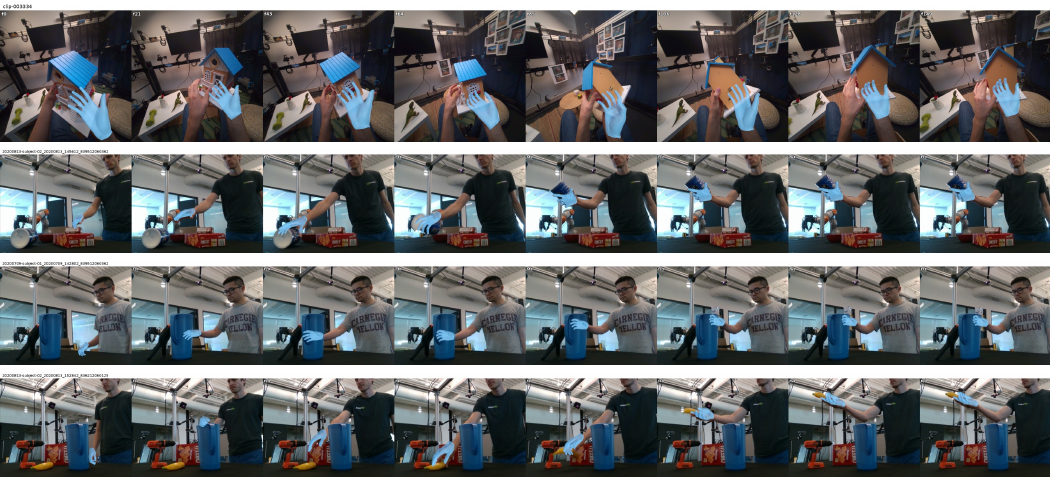}
\caption{HandFlow tracking on heavily occluded clips from DexYCB and HOT3D (eight sampled frames per sequence, overlaid on input views), with plausible poses inferred where the hand is largely unseen.}
\Description[HandFlow tracking under heavy occlusion]{HandFlow tracking on heavily occluded sequences from DexYCB and HOT3D, eight frames each overlaid on the input view, inferring plausible poses where the hand is unseen.}
\label{fig:tracking}
\end{figure*}

\noindent
\textbf{Occlusion analysis.}
We bin the test set by the fraction of invisible hand vertices into
25\%--50\%, 50\%--75\%, and 75\%--100\% occlusion. HandFlow leads on the full
set (3.88\,mm) and at light occlusion (25\%--50\%: 4.21\,mm), while
UniHand (also a generative method) edges ahead at heavier occlusion
(50\%--75\%: 4.25 vs.\ 4.35; 75\%--100\%: 4.26 vs.\ 4.85). Crucially, both
generative methods stay well below 5\,mm in the hardest 75\%--100\% bin,
whereas single-frame regressors rise to 5.68\,mm (WiLoR) and 7.37\,mm
(HandOccNet), confirming that generative modeling, rather than any specific
architecture, drives robustness under heavy occlusion. Deterministic
regressors tend to average over ambiguous observations under heavy occlusion,
whereas our generative formulation recovers coherent hand completions, as
Figure~\ref{fig:tracking} shows: HandFlow maintains plausible, kinematically
consistent poses even when the hand is largely unseen.
Figure~\ref{fig:dexycb-occ} further contrasts HandFlow with the baselines on
heavily occluded frames: the side and top-down views expose depth errors that
the overlay hides.

\subsection{World-Space Evaluation}
\label{sec:world_eval}

HOT3D is the more challenging benchmark: egocentric, multi-view captures with
frequent self- and object-occlusion, where we additionally evaluate the global
hand trajectory in world coordinates. Table~\ref{tab:hot3d} reports per-frame
pose accuracy (MPJPE, PA-MPJPE), world-space trajectory and shape accuracy
(W-MPJPE, WA-MPJPE), temporal smoothness (Accel), and root translation error
(RTE). Global camera pose is recovered via SLAM for all methods: HandFlow,
Dyn-HaMR, and the image-based baselines (denoted +SLAM) use
ViPE-SLAM~\cite{huang_vipe_2025}, while HaWoR uses its own SLAM. We discuss
world-space accuracy and temporal smoothness in turn.

\begin{figure*}[t]
\centering
\includegraphics[width=\linewidth]{Assets/f3.pdf}
\caption{HOT3D sequences rendered for HaMeR, HaWoR, HandFlow, and GT from a shared 3D viewpoint. HandFlow best matches GT in completeness and accuracy. Green denotes GT, blue HandFlow, orange HaWoR, and red HaMeR.}
\Description[HOT3D full-sequence comparison]{HOT3D motion sequences rendered from a shared 3D viewpoint for HaMeR, HaWoR, HandFlow, and GT, with HandFlow closest to GT.}
\label{fig:hot3d-seq}

\vspace{6pt}

\includegraphics[width=\linewidth]{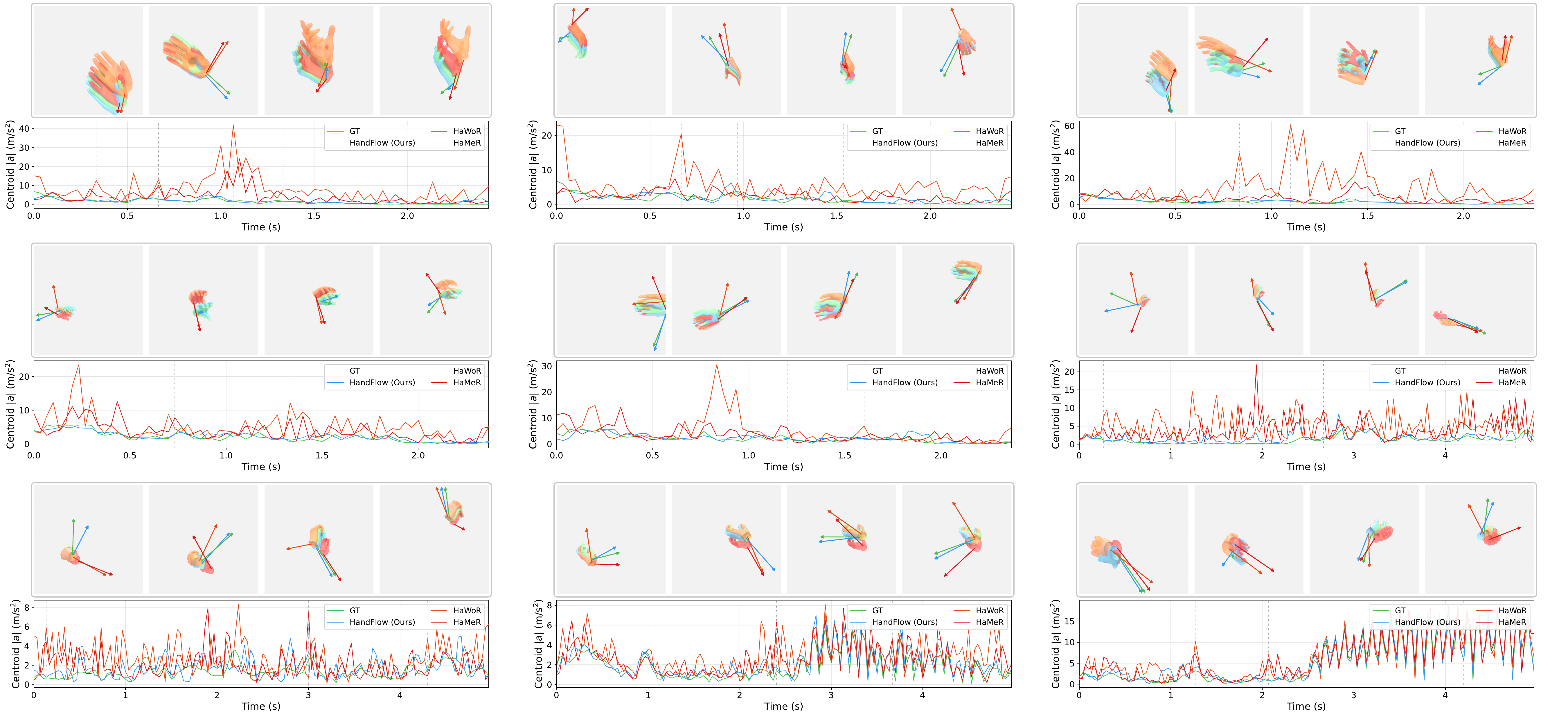}
\caption{For each sequence, four sampled frames show wrist acceleration arrows (length encodes magnitude), with a time-series plot of acceleration magnitude for GT and all methods beneath. HandFlow best matches GT in both magnitude and direction. Green denotes GT, blue HandFlow, orange HaWoR, and red HaMeR.}
\Description[Acceleration comparison]{Acceleration comparison: four wrist-arrow frames and an acceleration-magnitude time series per sequence for GT and all methods, where HandFlow best matches GT.}
\label{fig:accel}
\end{figure*}

\noindent
\textbf{World-space accuracy.}
HandFlow leads on every world-space and trajectory metric: W-MPJPE, WA-MPJPE,
and RTE, where camera-space pose error compounds with camera-pose drift. The
margin is largest in world space: WA-MPJPE is 16.17\,mm against 25.24 for
UniHand, 27.37 for HaWoR, and 31.01 for Dyn-HaMR, and RTE is 2.32\% against
the closest competitor's 2.44\%. Advantages in per-frame accuracy alone do not
translate into better global motion: UniHand attains the lowest per-frame
PA-MPJPE in the table (4.76\,mm), yet its world-space errors are roughly 50\%
higher than HandFlow's (W-MPJPE 63.97 vs.\ 43.00; WA-MPJPE 25.24 vs.\ 16.17).
This dissociation suggests that accurate global motion depends on temporal
coherence rather than single-frame precision alone.
Figure~\ref{fig:hot3d-seq} renders full sequences from a shared 3D viewpoint,
where HandFlow most closely matches the ground truth in completeness and
accuracy.

\noindent
\textbf{Temporal smoothness.}
HandFlow also achieves the lowest acceleration error (Accel 4.18\,m/s$^2$),
indicating temporally coherent reconstructions rather than per-frame estimates
strung together. Dyn-HaMR reaches the second-lowest Accel (4.77\,m/s$^2$), yet
its world-space errors remain far above HandFlow's (WA-MPJPE 31.01 vs.\
16.17), so its smoothness is bought at the expense of trajectory fidelity.
Figure~\ref{fig:accel} shows why: Dyn-HaMR's wrist-acceleration profile lags
the ground truth in phase during fast gestures and flattens genuine
high-frequency motion, whereas HandFlow tracks the ground truth in both
magnitude and direction, staying smooth without sacrificing fidelity.

This accuracy comes without the usual cost: reconstruction requires only a
single ODE integration per window, as quantified in
Section~\ref{sec:computational_efficiency}.

\subsection{Computational Efficiency}
\label{sec:computational_efficiency}

A practical hand pose estimator must deliver both accurate and timely reconstructions.
Table~\ref{tab:inference_time} reports inference time and resource consumption for all video-based methods on 150-frame HOT3D sequences (batch size 1, single NVIDIA H100-80\,GB GPU). In reconstruction time, HandFlow is over an order of magnitude faster than prior video-based methods, while simultaneously achieving the best accuracy among all compared methods.
The runtime breakdown reveals a structural shift: reconstruction accounts for only a small fraction of total latency, with SLAM dominating at 84\%.
In contrast, reconstruction (whether iterative optimization or multi-stage processing) dominates the latency of prior methods. Crucially, HandFlow and Dyn-HaMR share the same ViPE-SLAM frontend, making their 16$\times$ total-runtime gap a clean comparison of reconstruction speed. This inversion means HandFlow benefits directly from future SLAM acceleration, while baselines remain bottlenecked by their own pipelines.

\begin{table}[!t]
\centering
\caption{Computational efficiency comparison on 150-frame HOT3D sequences with batch size 1, measured on a single NVIDIA H100-80\,GB GPU. We report camera-space MPJPE (without world-coordinate alignment; see Table~\ref{tab:hot3d} for world-space results), reconstruction time, SLAM preprocessing time, total runtime, and peak GPU memory. HandFlow and Dyn-HaMR share the same ViPE-SLAM~\cite{huang_vipe_2025} frontend (hence identical SLAM time); HaWoR uses its native SLAM, so the reconstruction column isolates the methods' core difference. }
\label{tab:inference_time}
\small
\begin{adjustbox}{max width=\columnwidth}
\begin{tabular}{lccccc}
\toprule
Method & MPJPE $\downarrow$ & Recon. (s) & SLAM (s) & Total (s) & VRAM (GB) \\
\midrule
Dyn-HaMR & 58.12 & 288.58 & 16.21 & 304.79 & 1.1 \\
HaWoR & 68.11 & 38.13 & 15.78 & 53.91 & 6.9 \\
HandFlow & 23.79 & 3.19 & 16.21 & 19.40 & 13.1 \\
\bottomrule
\end{tabular}
\end{adjustbox}
\end{table}

HandFlow's higher VRAM stems from parallel denoising of all temporal windows for maximum throughput. Under memory constraints, window parallelism can be reduced with a modest throughput trade-off.
We further analyze the effect of ODE step count in Appendix~\ref{sec:app_ode_sweep}, and show that the framework is robust to the choice of frontend in Appendix~\ref{sec:app_frontend}.

\subsection{Ablation Studies}
\label{sec:ablation}

We ablate key design choices in HandFlow.
All variants are trained for 200 epochs, matching the full model.
Metrics are aggregated over the combined DexYCB (s0) and HOT3D (val) test sets, with the overlap ablation using a random subset for tractability.
Because this merged set differs from the per-benchmark splits of the main tables, absolute values are not comparable across tables, but relative ordering among variants remains informative.

\noindent
\textbf{Conditioning modalities.}
Table~\ref{tab:ablation} (top) removes each conditioning signal in turn. Dropping visual features causes a catastrophic drop (F@5 66.63\,$\to$\,13.42\%, PA-MPJPE 4.75\,$\to$\,8.47\,mm), confirming dense visual features are essential. Dropping the skeleton degrades translation and smoothness more moderately (MPJPE 18.73\,$\to$\,22.86\,mm), so sparse skeletal cues act as a complementary geometric anchor rather than the primary source of accuracy.

\noindent
\textbf{Continuous masking.}
For cmask (Eq.~\ref{eq:cmask}), random masking during training (zeroing $m_i$ with probability $p$) acts as a temporal regularizer whose effect splits along metric type (Table~\ref{tab:ablation}, middle). Per-frame accuracy favors \emph{no} masking ($p{=}0$ gives the lowest MPJPE), while temporal and global-coherence metrics favor masking: without it, each frame is predicted from its own observation with no pressure to exploit temporal context, yielding jagged trajectories. This yields a per-frame-vs-coherence trade-off governed by $p$: raising $p$ from 0 to 0.2 costs only 0.36\,mm of MPJPE but cuts acceleration error by 33\% and W-MPJPE by 2.28\,mm, while $p{=}0.5$ hurts accuracy further with no smoothness gain. We adopt $p{=}0.2$. Disabling confidence weighting at inference (w/o conf.) has a far larger effect: MPJPE rises by 86\% while PA-MPJPE is essentially unchanged, indicating that confidence-weighted conditioning mainly stabilizes global translation against noisy positional cues on low-confidence frames. 

\noindent
\textbf{Denoiser architecture.}
Table~\ref{tab:ablation} (bottom) compares the dual-stream denoiser against three parameter-matched variants. Replacing the dual-stream blocks with naive cross-attention, where condition tokens are read but cannot exchange information with latent tokens, degrades every metric substantially (MPJPE 18.73\,$\to$\,22.04\,mm; acceleration 4.08\,$\to$\,6.12\,m/s$^2$): a one-way read of the conditioning is insufficient, and bidirectional exchange between streams is essential. Among variants that preserve bidirectional fusion, Only Single-Stream stays closest to the full model on every metric (MPJPE 19.80 vs.\ 21.12\,mm for Only Dual-Stream), showing single-stream self-attention already fuses condition and latent tokens effectively. The full model combining both stages remains best, but its margin over Only Single-Stream is modest: the principal gain is the bidirectional fusion that naive cross-attention lacks, not the addition of the single-stream stage.

\begin{table}[!ht]
\centering
\caption{Ablation studies. All metrics are computed in camera space and aggregated over the full combined DexYCB (s0) and HOT3D (val) test set (the overlap ablation in Table~\ref{tab:overlap_fusion} uses a random subset thereof for tractability). All variants and the full model use a 200-epoch checkpoint, matching the main model's training budget. Architecture variants are parameter-matched to the full model within $\pm 5\%$. Our full model in \textbf{bold}. M. is short for MPJPE. $p$ denotes the training-time mask ratio (the full model uses $p{=}0.2$); \emph{w/o conf.} disables inference-time confidence weighting.}
\label{tab:ablation}
\small
\begin{adjustbox}{max width=\columnwidth}
\begin{tabular}{lccccccc}
\toprule
Setting & M. $\downarrow$ & PA-M. $\downarrow$ & W-M. $\downarrow$ & WA-M. $\downarrow$ & Accel $\downarrow$ & AUC $\uparrow$ & F@5 $\uparrow$ \\
\midrule
HandFlow (full) & \textbf{18.73} & \textbf{4.75} & \textbf{38.44} & \textbf{13.60} & \textbf{4.08} & \textbf{0.9111} & \textbf{66.63} \\
\midrule
\multicolumn{8}{l}{\textit{Conditioning modalities}} \\
\midrule
\;\; Image only    & 22.86 & 4.82 & 45.61 & 16.07 & 4.61 & 0.9098 & 65.67 \\
\;\; Skeleton only & 26.66 & 8.47 & 61.78 & 21.31 & 7.08 & 0.8288 & 13.42 \\
\midrule
\multicolumn{8}{l}{\textit{Continuous masking}} \\
\midrule
\;\; $p{=}0$   & 18.37 & 4.74 & 40.71 & 14.40 & 6.06 & 0.9113 & 66.39 \\
\;\; $p{=}0.5$ & 19.19 & 4.78 & 39.08 & 13.50 & 4.04 & 0.9105 & 66.83 \\
\;\; w/o conf. & 34.80 & 4.84 & 61.69 & 18.21 & 8.00 & 0.9091 & 64.84 \\
\midrule
\multicolumn{8}{l}{\textit{Denoiser architecture}} \\
\midrule
\;\; Naive Cross-Attention   & 22.04 & 6.03 & 50.69 & 16.58 & 6.12 & 0.8777 & 46.70 \\
\;\; Only Dual-Stream & 21.12 & 5.09 & 46.04 & 15.38 & 5.54 & 0.8932 & 55.95 \\
\;\; Only Single-Stream & 19.80 & 4.92 & 39.31 & 14.71 & 4.69 & 0.9036 & 61.14 \\
\bottomrule
\end{tabular}
\end{adjustbox}
\end{table}

\noindent
\textbf{Overlap inference.}
We adopt velocity-blended overlapping windows (Section~\ref{sec:vblend}) with overlap $o{=}2$ for all long-video experiments.
Velocity blending fuses predictions \emph{during} ODE integration, eliminating the boundary velocity jumps that post-hoc averaging introduces ($c_{>1}{=}18.9$ vs.\ $c_{1}{=}4.46$ at $o{=}2$), and improves accuracy over the no-overlap baseline.
The full fusion-strategy comparison is provided in Appendix~\ref{sec:app_overlap}.

\section{Conclusion}
\label{sec:conclusion}

We presented HandFlow, a fully generative flow-matching framework for monocular 4D hand reconstruction. By formulating reconstruction as conditional generation in the MANO parameter space, HandFlow moves beyond deterministic regression and learns to recover plausible and temporally coherent hand motion under ambiguous or unreliable observations. HandFlow reconstructs each window in a single ODE integration, keeping inference efficient without sacrificing accuracy or smoothness. Experiments on DexYCB and HOT3D show that HandFlow achieves state-of-the-art accuracy, temporal smoothness, and inference speed over prior image-based and video-based methods, suggesting that generative motion priors are a promising foundation for robust 4D hand recovery.

Three design choices underpin this behavior: a dual-stream transformer that jointly attends over latent hand parameters and visual-skeletal conditions across the full window, confidence-aware cmask that falls back on a learned motion prior when observations are unreliable, and velocity-blended overlapping-window inference that extends reconstruction to long videos along one continuous trajectory. Our results further show that accurate global motion stems from temporal coherence rather than single-frame precision.

% Bibliography
\bibliographystyle{ACM-Reference-Format}
\bibliography{ref/4D_HPE}

\clearpage

% --- Page 1: f1 (top) + f2 (bottom) ---

\clearpage
% --- Page 2: f3 (top) + f4 (bottom) ---

% Appendix
\clearpage
\appendix

\section{Evaluation Metrics}
\label{sec:app_metrics}

We adopt the following evaluation metrics:
\begin{itemize}
\item \textbf{MPJPE} measures the mean per-joint position error in millimeters without any geometric alignment. It is reported in camera space (Table~\ref{tab:inference_time}) and in world space after SLAM-based camera pose recovery (Table~\ref{tab:hot3d}); these two values are not directly comparable, as the latter additionally reflects camera-pose error.
\item \textbf{PA-MPJPE} applies Procrustes alignment (translation, rotation, and scale) before computing the error.
\item \textbf{RA-MPJPE} aligns only the root orientation.
\item \textbf{W-MPJPE} and \textbf{WA-MPJPE} differ from MPJPE only in their alignment. \textbf{W-MPJPE} aligns only the first two frames, so trajectory drift accumulates over the rest of the sequence and the metric directly reflects global trajectory accuracy. \textbf{WA-MPJPE} treats the entire sequence as a single point cloud and applies Procrustes alignment over all frames, which factors out drift and better reflects sequence-level shape consistency. As with MPJPE, both can be reported in either camera or world space.
\item \textbf{Acceleration error} measures the mean magnitude of the difference between predicted and ground-truth joint accelerations (m/s$^2$), reflecting temporal smoothness.
\item \textbf{RTE} (root translation error) measures the global translation error normalized by the total displacement of the hand trajectory after rigid alignment, reported in \% following~\cite{zhang_hawor_2025}.
\item \textbf{AUC} is the area under the MPJPE-threshold curve; we report $AUC_{RA}$ and $AUC_{PA}$, computed from RA-MPJPE and PA-MPJPE respectively.
\item \textbf{F@$v$} reports the fraction of frames with MPJPE below $v$~mm (we use $v{=}5$ and $v{=}15$).
\end{itemize}

For DexYCB, we report RA-MPJPE, PA-MPJPE, and their corresponding AUC values (Table~\ref{tab:dexycb_occlusion}); F@5 is additionally used in the ablation, and both F@5 and F@15 in the frontend-generality analysis.
For HOT3D, we report MPJPE, PA-MPJPE, W-MPJPE, WA-MPJPE, acceleration error, and RTE, where W-MPJPE and WA-MPJPE measure world-space accuracy.

\section{Implementation Details}
\label{sec:exp_details}

We use a hidden size of $D{=}512$, 16 attention heads, 8 dual-stream blocks followed by 16 single-stream blocks, and a window size of $T{=}16$ frames.
The model is trained with AdamW ($\beta_1{=}0.9, \beta_2{=}0.999$, weight decay $10^{-4}$), a learning rate of $10^{-4}$ with 5{,}000 warmup steps followed by cosine annealing, and a per-GPU batch size of 16 (effective batch 64 across 4 GPUs) for 200 epochs. All auxiliary losses are scaled by the mean sampled timestep $\bar{t}$.
The auxiliary loss weights are: $\mathcal{L}_\beta{=}1.0$, $\mathcal{L}_\text{vel}{=}0.5$, $\mathcal{L}_\text{acc}{=}0.5$, $\mathcal{L}_\text{reproj}{=}0.5$, $\mathcal{L}_\text{j3d}{=}0.05$.

\noindent
\textbf{Training data.}
HandFlow is trained on the training splits of DexYCB and HOT3D, the same two benchmarks used for evaluation, with the evaluation sequences held out.
For DexYCB we adopt the standard s0 protocol: training uses all 10 subjects and 8 camera views over 80\% of the sequences, while the s0 evaluation set (subjects~1--2, every 5th sequence) is excluded from training.
For HOT3D, whose official validation clips lack public ground truth, we split the official training set ourselves: within each participant, sequences are ranked alphabetically and every seventh sequence is held out for validation (13 sequences, 169 clips, ${\sim}11\%$ of the training clips), with the remaining 1{,}347 clips used for training. All re-evaluated baselines use this same split.
Following prior work~\cite{pavlakos_reconstructing_2024}, every hand is detected and cropped by HaMeR into a $256{\times}256$ right-hand region; MANO parameters are stored in axis-angle form and grouped into windows of $T{=}16$ frames, discarding windows with fewer than 50\% valid frames.
Full dataset-construction and pre-processing scripts will be released.

All models are trained on 4$\times$ NVIDIA H100-80\,GB GPUs. Training HandFlow takes approximately 24 hours for 200 epochs. For the computational efficiency evaluation (Table~\ref{tab:inference_time}), we average over 50 validation sequences due to the high inference cost of baseline methods.

\section{ODE Steps Sweep Analysis}
\label{sec:app_ode_sweep}

\begin{figure}[ht]
\centering
\includegraphics[width=0.85\linewidth]{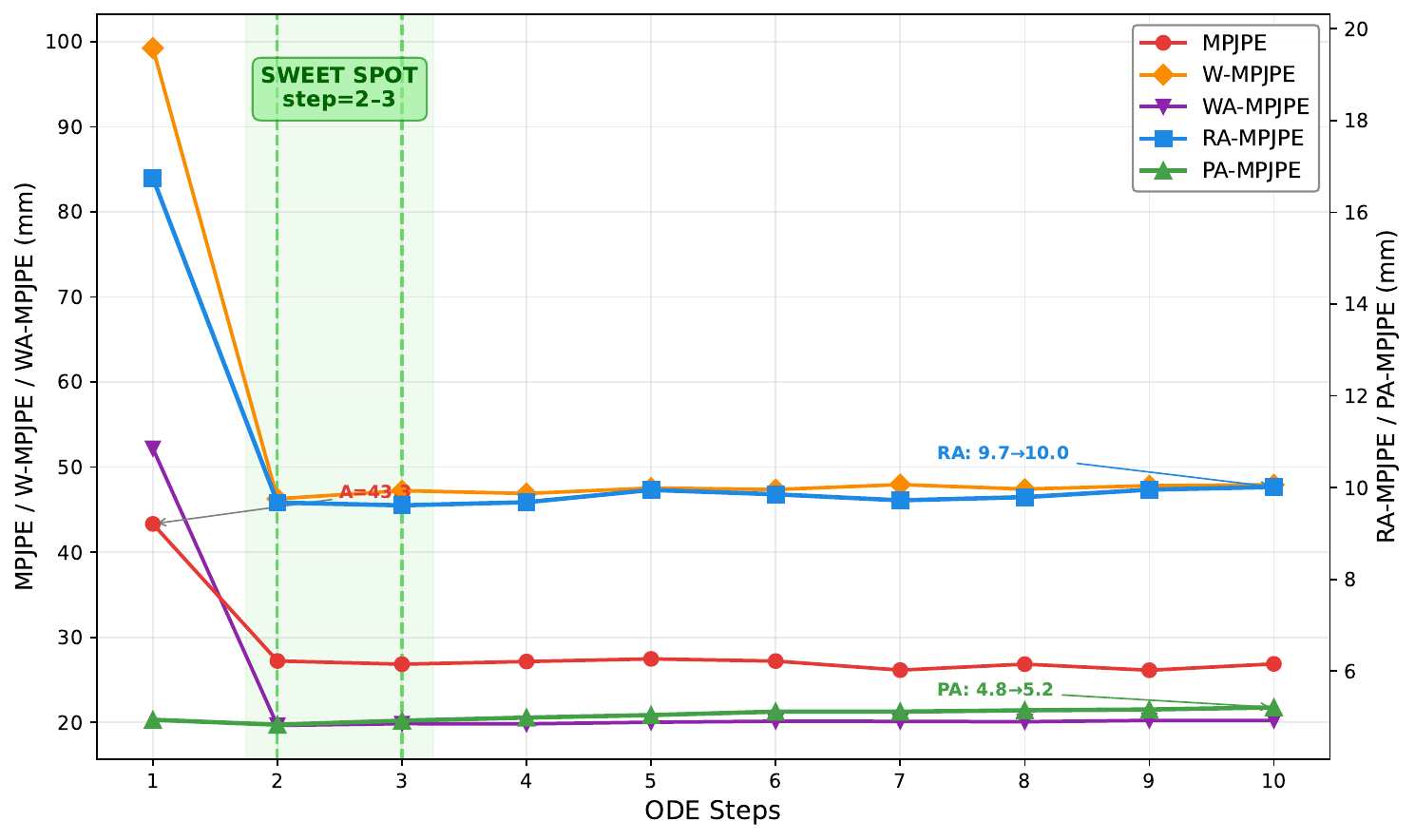}
\caption{
\textbf{ODE steps vs. MPJPE.} Five MPJPE variants (M, PA-M, RA-M, W-M, WA-M) across 1--10 Euler integration steps.
}
\Description[ODE steps vs. MPJPE plot]{Line plot showing five MPJPE variants (M, PA-M, RA-M, W-M, WA-M) across 1--10 Euler integration steps. Step 1 has high error (~43 mm); from step 2 onward all metrics plateau.}
\label{fig:ode_sweet_spot}
\end{figure}

We analyze the effect of varying the number of ODE integration steps during inference.
Using a fixed model checkpoint and Euler solver, we sweep steps from 1 to 10 on the full validation set (DexYCB s0 + HOT3D val clips, 42,384 frames total).
Figure~\ref{fig:ode_sweet_spot} reports five MPJPE variants across step counts.

\textbf{Key observations.}
(1)~Step~1 collapses under Euler truncation error (MPJPE~$\approx$~43~mm), as a single-step integration cannot resolve the continuous flow trajectory.
(2)~From step~2 onward, performance reaches a stable plateau: MPJPE drops sharply from $\approx$43~mm at step~1 to $\approx$27~mm at step~2 and then stays largely flat. PA-MPJPE remains nearly constant across the sweep (4.83--5.21~mm), confirming that per-frame pose quality is largely step-count-independent. We adopt step~3 rather than step~2, as the latter sits at the edge of the plateau and offers a thinner safety margin against distribution shift.

\section{Noise Sensitivity Analysis}
\label{sec:app_noise}

HandFlow is fully generative: at inference it starts from random initial noise $\bar{\mathbf{x}}_0 \sim \mathcal{N}(\mathbf{0}, \mathbf{I})$ and integrates the flow ODE deterministically with the Euler solver (Section~\ref{sec:flow_matching}). Given $\bar{\mathbf{x}}_0$, the entire integration is fixed, so the only stochastic component at inference is the initial noise itself. A natural concern is whether the recovered hand states depend on this random draw. To quantify this, we run 10 independent inference passes with different initial noise samples (random seeds 0--9) on the full combined DexYCB (s0) and HOT3D (val) validation set (482 sequences, 42{,}384 frames), using the same fixed checkpoint and 3-step Euler setting as our reported results (cf.\ Section~\ref{sec:app_ode_sweep}). Table~\ref{tab:noise_sensitivity} reports the mean, standard deviation, range, and coefficient of variation (CV) of each metric across the 10 runs.

% 噪声敏感性分析表格：10-seed sweep 的离散度统计（combined DexYCB s0 + HOT3D val）
\begin{table}[!ht]
\centering
\caption{
\textbf{Sensitivity to the initial noise.}
We run 10 independent inferences with different initial noise samples (random seeds 0--9)
on the full combined DexYCB (s0) and HOT3D (val) validation set (482 sequences, 42{,}384 frames),
using the fixed checkpoint and 3-step Euler setting of our reported results.
For each metric we report the mean, standard deviation, range (max--min), and coefficient
of variation (CV $=$ std/mean, in \%) across the 10 runs.
MPJPE variants are in mm and acceleration error in m/s$^2$.
The combined mean is not directly comparable to the per-benchmark
Tables~\ref{tab:dexycb_occlusion} and~\ref{tab:hot3d}; only the dispersion
(std, range, CV) is meaningful here.
Rows are ordered by CV; lower CV indicates lower sensitivity to the random initial noise.
}
\label{tab:noise_sensitivity}
\small
\begin{adjustbox}{max width=\columnwidth}
\begin{tabular}{lcccc}
\toprule
Metric & mean & std & range & CV (\%) $\downarrow$ \\
\midrule
PA-MPJPE  & 4.76  & 0.008 & 0.025 & 0.16 \\
RA-MPJPE  & 10.30 & 0.032 & 0.104 & 0.31 \\
WA-MPJPE  & 13.46 & 0.072 & 0.222 & 0.53 \\
MPJPE     & 18.51 & 0.133 & 0.382 & 0.72 \\
W-MPJPE   & 38.02 & 0.312 & 0.779 & 0.82 \\
Accel     & 3.98  & 0.045 & 0.148 & 1.12 \\
\bottomrule
\end{tabular}
\end{adjustbox}
\end{table}

\noindent
\textbf{Pose accuracy is essentially noise-independent.}
PA-MPJPE varies by only 0.025~mm across the 10 noise samples (CV 0.16\%), and RA-MPJPE by 0.104~mm (CV 0.31\%). Different initial noise samples thus converge to nearly identical solutions, indicating that the rectified-flow trajectory is sufficiently straight that the final prediction depends only weakly on the initial noise. This empirically corroborates the straight-path assumption underlying rectified flow~\cite{liu_flow_2023}.

\noindent
\textbf{Global translation is slightly more sensitive, yet within 1\%.}
Metrics that include the global translation, MPJPE (CV 0.72\%) and W-MPJPE (CV 0.82\%), vary roughly an order of magnitude more than pose-aligned PA-MPJPE. This reflects that absolute global placement is harder to pin down than articulated pose, which is consistent with translation being the least constrained component of the MANO state. Still, every spatial CV remains below 1\%.

\noindent
\textbf{Temporal-metric CVs are inflated by small denominators.}
Acceleration error shows a higher CV (1.12\%), and RTE (1.75\%, not tabulated) higher still, but their absolute ranges are tiny (0.15~m/s$^2$ and 0.09\%); the large CV is a statistical artifact of dividing a small standard deviation by an even smaller mean. A practical consequence is that the improvements over baselines reported in Tables~\ref{tab:dexycb_occlusion} and~\ref{tab:hot3d}, on the order of millimeters, are orders of magnitude larger than the seed-induced spread and thus reflect genuine method differences rather than noise in the random initialization.

Together with the ODE-step sweep (Section~\ref{sec:app_ode_sweep}), this analysis shows that HandFlow's inference is robust along two independent axes: the solver budget and the random starting point.

\section{Frontend Generality Analysis}
\label{sec:app_frontend}

By default, HandFlow is conditioned on HaMeR~\cite{pavlakos_reconstructing_2024}, a single hand-specialized network (ViT-Huge backbone, $\sim$600M parameters, pre-trained on large-scale hand data) that jointly provides both the dense visual features and the 21 2D keypoints.
To probe robustness to the frontend, we replace this entire module with off-the-shelf components that are \emph{not} specialized for hands and are strictly weaker on every axis:
DINOv3-Base~\cite{simeoni_dinov3_2025} ($\sim$86M, a general-purpose ViT self-supervised on natural images with no hand-specific pre-training) supplies the visual features, and a MediaPipe~\cite{lugaresi_mediapipe_2019} hand detector ($\sim$2M) supplies the 2D keypoints.
The combined frontend is $\sim$88M parameters (roughly $1/7$ of HaMeR), and its two components are trained independently rather than jointly.
All other components (the frame compressor, skeleton projection, cmask mechanism, dual-stream denoiser, and loss functions) remain unchanged.

Table~\ref{tab:frontend_generality} reports the results on both benchmarks.
Despite the strictly weaker frontend, HandFlow degrades gracefully: the drop is moderate and consistent across metrics (PA-MPJPE $+15\%$ on DexYCB and $+16\%$ on HOT3D, MPJPE $+18\%$ on HOT3D).
More importantly, even with this generic frontend, HandFlow still outperforms strong video-based baselines (cf.\ Tables~\ref{tab:hot3d} and~\ref{tab:dexycb_occlusion}): on DexYCB, PA-MPJPE 4.46\,mm vs.\ HaWoR 4.76\,mm; on HOT3D, W-MPJPE 57.28\,mm vs.\ Dyn-HaMR 69.11\,mm.
Because HaMeR couples its backbone and keypoint head in a single network, we treat the frontend as one replaceable module and do not further decompose the residual drop.
These results indicate that HandFlow's accuracy is \emph{not contingent on a hand-specialized 600M backbone}: the rectified-flow formulation and dual-stream conditioning provide most of the performance, and the framework degrades only modestly under a generic frontend that is both $\sim$7$\times$ smaller and free of hand-specific pre-training.

% 前端通用性分析表格：HaMeR 前端 vs 通用 DINOv3+MediaPipe 前端
\begin{table}[t]
\centering
\caption{
\textbf{Frontend generality.} HandFlow accuracy when the HaMeR frontend is replaced by a generic DINOv3-Base backbone and MediaPipe skeleton detector (cf.\ Tables~\ref{tab:hot3d} and~\ref{tab:dexycb_occlusion} for specialized baselines).
}
\label{tab:frontend_generality}
\resizebox{\columnwidth}{!}{%
\begin{tabular}{lccccccccc}
\toprule
& & \multicolumn{4}{c}{HOT3D} & \multicolumn{4}{c}{DexYCB (s0)} \\
\cmidrule(lr){3-6} \cmidrule(lr){7-10}
Frontend & Params & M$\downarrow$ & PA$\downarrow$ & W-M$\downarrow$ & Accel$\downarrow$ & PA$\downarrow$ & AUC$\uparrow$ & F@5$\uparrow$ & F@15$\uparrow$ \\
\midrule
HaMeR (ViT-H) + HaMeR Skel. & $\sim$600M
  & \textbf{72.00} & \textbf{5.49} & \textbf{43.00} & \textbf{4.18}
  & \textbf{3.88} & \textbf{0.922} & \textbf{80.30} & \textbf{99.82} \\
DINOv3-Base + MediaPipe & $\sim$88M
  & 85.35 & 6.37 & 57.28 & 5.61
  & 4.46 & 0.917 & 75.81 & 97.20 \\
\bottomrule
\end{tabular}%
}
\end{table}

\section{Overlap Inference Analysis}
\label{sec:app_overlap}

We analyze the overlapping-window inference strategy of Section~\ref{sec:vblend}, comparing fusion methods across overlap sizes $o \in \{0,1,2,4\}$ on a random subset of the combined DexYCB (s0) and HOT3D (val) test set.
We decompose acceleration error into $c_{1}$ (frames covered by a single window) and $c_{>1}$ (frames in overlap regions); Table~\ref{tab:overlap_fusion} reports the results.

\begin{table}[!ht]
\centering
\caption{
\textbf{Overlap inference analysis.}
We compare fusion strategies across overlap sizes $o \in \{0,1,2,4\}$ on a
random subset of the combined DexYCB (s0) and HOT3D (val) test set;
all metrics are computed over this same mixed subset.
$c_{1}$ and $c_{>1}$ denote acceleration error for frames covered by one
and more-than-one window, respectively.
Absolute values are not directly comparable to the single-benchmark
Tables~\ref{tab:dexycb_occlusion} and~\ref{tab:hot3d}, but the comparison
across fusion strategies is fair.
Velocity blending eliminates boundary artifacts ($c_1 \approx c_{>1}$) and
improves reconstruction accuracy over the no-overlap baseline.
Best per column in \textbf{bold}. Selected configuration ($o{=}2$) highlighted in light green.
}
\label{tab:overlap_fusion}
\small
\begin{adjustbox}{max width=\columnwidth}
\begin{tabular}{lcccccc}
\toprule
Setting & $o$ & Accel $\downarrow$ & $c_1$ & $c_{>1}$ & RA-MPJPE $\downarrow$ & WA-MPJPE $\downarrow$\\
\midrule
No overlap & 0 & 4.49 & 4.49 & -- & 8.24 & 12.76\\
\midrule
\multirow{3}{*}{Post-hoc avg}
  & 1 & 5.23 & 4.48 & 12.8 & 8.28 & 13.37\\
  & 2 & 6.74 & 4.46 & 18.9 & 8.38 & 13.59\\
  & 4 & 9.12 & 4.45 & 26.1 & 8.55 & 17.19\\
\midrule
\multirow{3}{*}{\textbf{Velocity blend (ours)}}
  & 1 & 4.35 & 4.30 & \textbf{5.18} & 8.19 & 13.23 \\
  & \cellcolor{green!12}{2} & \cellcolor{green!12}{4.26} & \cellcolor{green!12}{4.19} & \cellcolor{green!12}{5.43} & \cellcolor{green!12}{\textbf{8.13}} & \cellcolor{green!12}{12.30}\\
  & 4 & \textbf{4.18} & \textbf{4.08} & 5.86 & 8.16 & \textbf{11.97} \\
\bottomrule
\end{tabular}
\end{adjustbox}
\end{table}

\noindent
\textbf{Post-hoc averaging introduces boundary artifacts.}
Running an independent ODE per window and averaging the final states of overlapping frames concentrates error at overlap boundaries ($c_{>1}{=}18.9$ vs.\ $c_{1}{=}4.46$ at $o{=}2$): independently evolved ODE trajectories diverge, and averaging their endpoints introduces non-physical velocity jumps.

\noindent
\textbf{Velocity blending fuses during integration.}
By maintaining a single shared state per global frame and blending velocities at each ODE step (Section~\ref{sec:vblend}), velocity blending achieves the lowest overall acceleration error and keeps boundary-frame error close to interior-frame levels ($c_{1} \approx c_{>1}$), confirming that per-step blending prevents trajectory divergence rather than smoothing its aftereffects.
Velocity blending with $o{>}0$ further improves over the no-overlap baseline in both acceleration error and RA-MPJPE, showing that overlapping observations provide complementary information: frames near window edges benefit from the richer temporal context of adjacent windows.
We adopt $o{=}2$, which captures most of the benefit of larger overlaps at lower window-count overhead.

\section{Limitations}

First, the \emph{cmask} mechanism conflates hand occlusion with the hand leaving the camera field of view, since both manifest as low HaMeR detection confidence.
These scenarios differ fundamentally: under partial occlusion the generative prior can plausibly complete missing information, but when the hand is fully out of frame there is no visual evidence to condition on and predictions degrade substantially, yet the model cannot detect or flag this failure mode.
Second, reconstruction quality depends on the accuracy of camera intrinsic parameters.
When intrinsics are not known \emph{a priori} and must be estimated online via ViPE, sequences with limited camera motion provide insufficient geometric constraints for accurate estimation, leading to significant degradation in reconstruction quality.

Third, because HandFlow operates exclusively in the MANO parameter space rather than a dense surface representation, its accuracy is bounded both by the limited expressiveness of the MANO model and by the precision of the MANO-format annotations, which are themselves a lossy approximation of the true hand geometry.

\end{document}